\documentclass{article}

\usepackage[preprint]{neurips_2025}

\usepackage[utf8]{inputenc} 
\usepackage[T1]{fontenc}    
\usepackage{hyperref}       
\usepackage{url}            
\usepackage{booktabs}       
\usepackage{amsfonts}       
\usepackage{nicefrac}       
\usepackage{microtype}      
\usepackage{xcolor}         
\usepackage{graphicx}
\usepackage{multicol}
\usepackage{multirow}
\usepackage{subcaption} 
\usepackage{soul}

\title{Multiple Weaks Win Single Strong: Large Language Models Ensemble Weak Reinforcement Learning Agents into a Supreme One}

\author{%
    Yiwen Song\footnotemark[1], Qianyue Hao\footnotemark[1], Qingmin Liao, Jian Yuan, Yong Li\footnotemark[2]\\
    Department of Electronic Engineering, BNRist, Tsinghua University \\
    Beijing China
}

\begin{document}
\footnotetext[1]{The two authors contribute equally to this work.}
\footnotetext[2]{Corresponding author, email: \texttt{liyong07@tsinghua.edu.cn}}
\renewcommand{\thefootnote}{\arabic{footnote}}
\maketitle

\begin{abstract}
Model ensemble is a useful approach in reinforcement learning (RL) for training effective agents.
Despite wide success of RL, training effective agents remains difficult due to the multitude of factors requiring careful tuning, such as algorithm selection, hyperparameter settings, and even random seed choices, all of which can significantly influence an agent’s performance.
Model ensemble helps overcome this challenge by combining multiple weak agents into a single, more powerful one, enhancing overall performance.
However, existing ensemble methods, such as majority voting and Boltzmann addition, are designed as fixed strategies and lack a semantic understanding of specific tasks, limiting their adaptability and effectiveness.
To address this, we propose \textbf{LLM-Ens}, a novel approach that enhances RL model ensemble with task-specific semantic understandings driven by large language models (LLMs).
Given a task, we first design an LLM to categorize states in this task into distinct “situations”, incorporating high-level descriptions of the task conditions.
Then, we statistically analyze the strengths and weaknesses of each individual agent to be used in the ensemble in each situation.
During the inference time, LLM-Ens dynamically identifies the changing task situation and switches to the agent that performs best in the current situation, ensuring dynamic model selection in the evolving task condition.
Our approach is designed to be compatible with agents trained with different random seeds, hyperparameter settings, and various RL algorithms.
Extensive experiments on the Atari benchmark show that LLM-Ens significantly improves the RL model ensemble, surpassing well-known baselines by up to 20.9\%.
For reproducibility, our code is open-source at \url{https://anonymous.4open.science/r/LLM4RLensemble-F7EE}.
\end{abstract}

\section{Introduction}
\label{introduction}


Reinforcement learning (RL) has made unprecedented progress and been recognized as an effective tool for training intelligent agents in solving sequential decision-making tasks~\cite{sutton2018reinforcement,franccois2018introduction}.
Particularly, deep RL has achieved remarkable success in highly complicated tasks, including real-time games~\cite{silver2017mastering,vinyals2019grandmaster,berner2019dota,ye2021mastering}, chip design~\cite{mirhoseini2021graph}, smart city governance~\cite{hao2021hierarchical,hao2022reinforcement,hao2023gat,zheng2023spatial,zheng2024survey,wang2024dyps,wang2025coopride}, where the RL agents are increasingly exhibiting performance surpassing human professionals nowadays.
In a wide range of RL research, model ensemble is a useful approach for training effective RL agents.
Because a large number of factors will influence the agent’s performance in RL training, it is difficult to consistently obtain reliable RL agents, especially when scaling to more complex tasks.
While the choice of RL algorithm is crucial, hyperparameter settings within a single algorithm~\cite{adkins2024method}, such as learning rate and discount factor, can significantly impact the outcome and often require careful tuning and extensive trial-and-error.
Moreover, even merely diverse random seeds can lead to completely different outcomes~\cite{colas2018many}.

Similar to ensemble methods in machine learning~\cite{ganaie2022ensemble,dong2020survey}, RL can benefit from combining multiple weak agents, which may individually require minimal hyperparameter tuning and exhibit suboptimal performance in isolation, to produce a powerful agent outperforming any single weak one.
This strategy leverages the diversity of the agents to achieve better generalization and more stable performance across different tasks~\cite{song2023ensemble}.
However, existing ensemble methods, including majority voting~\cite{fausser2015neural}, weighted aggregation~\cite{perepu2020reinforcement}, and Boltzmann addition~\cite{wiering2008ensemble}, typically rely on fixed, rule-based strategies to combine the agents' actions. These approaches do not account for the specific characteristics of the task at hand, lacking a semantic understanding of the task’s context.
As a result, such methods fail to fully exploit the strengths of each agent to adapt to the complex environments.

There exist several challenges in addressing such limitation.
First, different tasks have unique characteristics that require tailored ensemble strategies to fully exploit the strengths of each agent.
However, RL tasks span a wide range of environments, and designing rule-based strategies for each possible scenario is neither practical nor efficient.
Second, RL tasks often involve multiple steps where the state continuously evolving over time.
This dynamic nature makes it difficult to rely on pre-defined ensemble strategies, as they may not adapt well to the changing status during the task.

Facing these challenges, we propose \textbf{LLM-Ens}, a framework that leverages the contextual understanding and reasoning abilities of large language models (LLMs) to enhance RL model ensemble.
The process unfolds in three stages.
First, for a given task, we design an LLM to categorize the possible states into distinct “situations” based on the task information, which provide generalized descriptions of the task-specific conditions with semantic understanding, enhancing the adaptability across different tasks.
Next, we statistically analyze the performance of each agent that will be used in the ensemble in each categorized situation, identifying their strengths and weaknesses.
Finally, during inference, LLM-Ens dynamically adapts to the evolving task by identifying the current situation and selecting the agent that performs best under the identified situation, ensuring the ensemble can respond to changes in the environment flexibly.
We conduct extensive experiments on the Atari benchmark~\cite{bellemare2013arcade,kaiser2019model} to evaluate the performance of LLM-Ens.
The results demonstrate that LLM-Ens, through model ensemble, effectively combines multiple weak agents to create a supreme one with optimal performance, outperforming existing RL model ensemble baselines.
Our approach is designed to work seamlessly with agents trained using different random seeds, hyperparameter settings, and RL algorithms, maximizing its potential applications.

In summary, the main contributions of this work include:
\begin{itemize}
    \item We propose LLM-Ens, a framework that leverages LLMs to generate adaptive and flexible RL model ensemble strategies in different tasks, which effectively combines multiple weak agents into a supreme one with optimal performance.
    \item We conduct extensive experiments to verify the ability of our method to improve RL model ensemble across various RL tasks. The results indicate that our method surpasses established baselines by up to 20.9\%.
    \item Our method is compatible with agents trained using different random seeds, hyperparameter configurations, and RL algorithms, ensuring broad applicability across diverse settings.
\end{itemize}
\section{Problem Formulation}

\subsection{Markov Decision Process (MDP)}
Markov Decision Process (MDP) is the fundamental framework for reinforcement learning (RL), providing a mathematical structure for modeling sequential decision-making problems where an agent interacts with a dynamic environment. The goal of the agent is to learn an optimal policy that maximizes cumulative rewards over time by making a series of decisions based on observed states.

Formally, an MDP is defined as a quintuple $(\mathcal{S}, \rho, \mathcal{A}, P, R)$, where $\mathcal{S}$ denotes the set of all possible states that describe the environment, and $\rho \in \Delta(\mathcal{S})$ represents the probability distribution over the initial states. Here, $\Delta(\mathcal{S})$ refers to the set of all probability distributions defined over $\mathcal{S}$. The action space, denoted as $\mathcal{A}$, includes all possible actions that the agent can take at any given state. The transition dynamics of the environment are modeled by a probability function $P: \mathcal{S} \times \mathcal{A} \to \Delta(\mathcal{S})$, which defines the likelihood of transitioning to a new state $s' \in \mathcal{S}$ given the current state $s \in \mathcal{S}$ and action $a \in \mathcal{A}$. Additionally, the reward function $R: \mathcal{S} \times \mathcal{A} \to \mathbb{R}$ specifies the immediate reward obtained when the agent executes action $a$ in state $s$.

At each discrete time step $t$, the agent observes the current state $s_t \in \mathcal{S}$, selects an action $a_t \in \mathcal{A}$ according to its policy, receives a scalar reward $r_t = R(s_t, a_t)$, and transitions to the next state $s_{t+1} \sim P(s_{t+1} | s_t, a_t)$. The agent’s objective in an MDP is to maximize the expected cumulative reward over time. This is typically defined using a discounted return, given by $G_t = \sum_{k=0}^{\infty} \gamma^k r_{t+k}$, where $\gamma \in [0,1]$ is the discount factor, which determines the relative importance of future rewards compared to immediate rewards. A lower value of $\gamma$ makes the agent more short-sighted, favoring immediate rewards, while a higher value encourages long-term planning.

\subsection{Large Language Models (LLMs)}
Large Language Models (LLMs) are advanced deep neural networks trained on vast corpora of text data, primarily designed to predict the probability distribution of the next token in a sequence. These models, often built using the transformer architecture, excel at capturing complex language patterns and syntactic structures by leveraging self-attention mechanisms. Given an input sequence $\{w_1, w_2, ..., w_{t-1}\}$, an LLM predicts the next word $w_t$ by maximizing the conditional probability:

\begin{equation}
    \prod_{t=1}^{T} P(w_t | w_1, w_2, ..., w_{t-1}).
\end{equation}

Recent years have witnessed remarkable advancements in LLMs, leading to the development of highly capable models such as the GPT series~\cite{Brown2020gpt3,kalyan2023survey,achiam2023gpt,zhong2024evaluation,openai2024o3}, the Llama family~\cite{touvron2023llama,dubey2024llama}, and the PaLM series~\cite{chowdhery2023palm}. These models have demonstrated impressive performance across a wide array of natural language processing (NLP) tasks, including text generation, translation, summarization~\cite{hao2024hlm,lan2024stance}, question answering~\cite{zhao2023survey,chang2024survey}, and complex reasoning tasks~\cite{xu2025towards}.
\begin{figure*}[ht]
    \begin{center}    
    \includegraphics[width=0.9\linewidth]{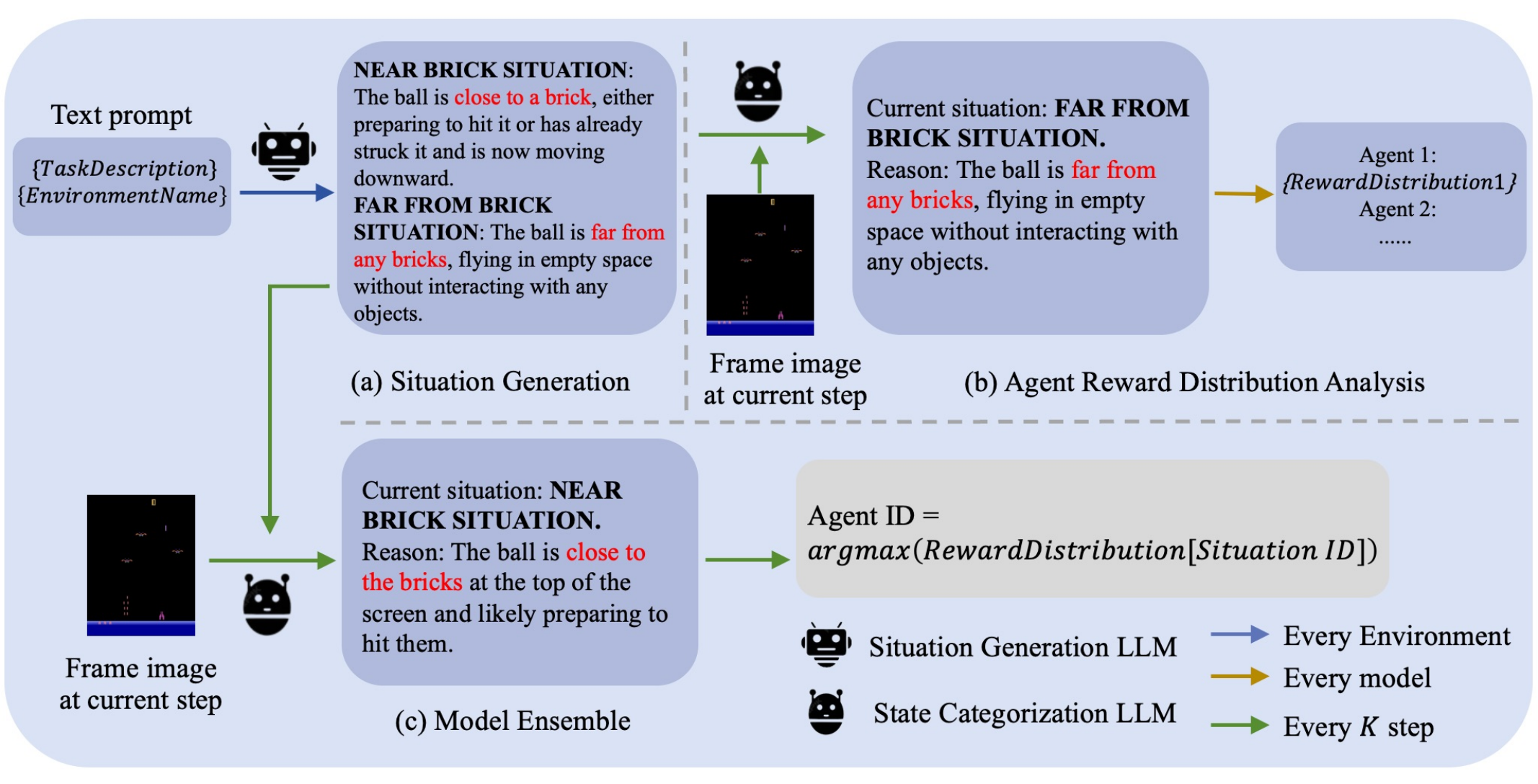}
    \caption{Illustration of LLM-Ens, which leverages LLMs to dynamically categorize task-specific situations and adaptively select the most suitable agent, enhancing RL model ensemble performance.}
    \label{fig1}
    \end{center}
\end{figure*}

\section{Methods}

\subsection{Overview}
In this paper, we propose a novel framework that utilizes LLMs to enhance RL model ensemble by dynamically adapting to task-specific situations, as illustrated in Figure~\ref{fig1}.
The aim of RL model ensemble is to combine several weak RL agents into a strong one that outperforms any of the orignal weak agents.
The framework consists of three stages. 
In the first stage, Situation Generation, the Situation Generation LLM categorizes all possible states into a set of generalized situations based on a detailed task description, providing a structured representation of the environment (Section~\ref{method1}). 
In the second stage, Agent Reward Distribution Analysis, the State Categorization LLM identifies the current situation every $K$ step and computes the average reward achieved by each agent in each situation, creating a reward distribution that reflects agents' performance across all situations (Section~\ref{method2}). 
Here, $K$ is the hyperparameter that defines the interval at which the State Categorization LLM is used. 
In the final stage, Model Ensemble, we use the same State Categorization LLM to identify the current situation every $K$ step in real time.
Then, we select the agent with the highest average reward in the currently identified situation, ensuring that actions are generated by the most suitable agent (Section~\ref{method3}). 

This structured process leverages the complementary strengths of multiple agents, enabling a dynamic and context-aware approach to RL model ensemble.

\subsection{Situation Generation}
\label{method1}
To effectively evaluate the performance of different agents across various situations, we design the Situation Generation LLM to categorize all potential states of a specific task into several general classes, referred to as situations. 
To achieve this, we begin by outlining the fundamental components of the task as \textit{\{TaskDescription\}} in the prompt:

\textbf{Task Description}: The task is a reinforcement learning problem where an agent \textit{\{TaskDetails\}}. The action space is \textit{\{ActionDetails\}}. The agent receives a reward of \textit{\{RewardDetails\}}. The game ends when \textit{\{EndConditions\}}. The goal is to \textit{\{GoalDetails\}}.

This ensures that the LLM comprehends the task's specific characteristics, which enables it to perform the state classification.
Next, we construct a customized prompt for the situation generation LLM as:

\textbf{Situation Generation Prompt}: You are classifying all states in the Atari \textit{\{EnvironmentName\}} environment into a few situations. \textit{\{TaskDescription\}}. Please provide your classification and a brief description of it. Only present the classification method you consider most reasonable, using as few categories as possible. \textit{\{OutputFormat1\}}.

This prompt equips the Situation Generation LLM with the required context to categorize all possible states in the environment into several generalized situations.
Furthermore, it requires the LLM to provide concise descriptions of each situation, with the goal of supplying more information for the subsequent classification of real-time state observations into the appropriate situation. 
The \textit{\{OutputFormat1\}} for Situation Generation is:

\textbf{Output Format 1 (Situation Generation)}:  Your output format should be: \{[situation 1]: [description 1], [situation 2]: [description 2], ...\}.

\subsection{Agent Reward Distribution Analysis}
\label{method2}
To monitor real-time interactions between the agent and the environment during inference, we design the State Categorization LLM in to determine which situation the current state belongs to. The classification of the current state is performed based on the \textit{\{GeneratedSituations\}} defined during the Situation Generation phase. This step ensures that the framework continuously adapts to changes in the task's dynamics by accurately identifying the state situation at regular intervals. The situation ID is generated by instructing the state categorization LLM with a structured prompt as:

\textbf{State Categorization Prompt}: In the Atari \textit{\{EnvironmentName\}} environment, many different states may occur. \textit{\{TaskDescription\}}. The states faced by the agent can be divided into \textit{\{SituationNum\}} general categories, which are listed as follows: \textit{\{GeneratedSituations\}}. Please classify the input image into one of these situations and attach a brief reason for your conclusion. \textit{\{OutputFormat2\}}.

The \textit{\{OutputFormat2\}} for State Categorization is:

\textbf{Output Format 2 (State Categorization)}: 
Use the output format: \{[situation ID], [reason]\}.

Using this prompt, the State Categorization LLM periodically analyzes the real-time interaction between the agent and the environment, updating the current situation ID accordingly. The categorization is performed every $K$ steps, where $K$ is a tunable hyperparameter that defines the interval at which the state categorization is updated. This approach ensures that the framework maintains a dynamic understanding of the task conditions while balancing computational efficiency.

After determining the situation ID, we compute the average reward achieved by each agent within that specific situation. This reward computation involves aggregating the rewards from all interactions associated with the identified situation and calculating the mean reward for each model. The result is a comprehensive \textit{\{RewardDistribution\}}, which summarizes the performance of all models across all situations. Mathematically, the average reward for agent $m$ in situation $s$ is calculated as:
\[
R_{m,s} = \frac{1}{N_s} \sum_{i=1}^{N_s} r_{i,m}
\]
where $R_{m,s}$ is the average reward for agent $m$ in situation $s$, $N_s$ is the total number of occurrences of situation $s$, and $r_{i,m}$ is the reward received by agent $m$ during the $i$-th occurrence of situation $s$.

The \textit{\{RewardDistribution\}} provides critical insights into the strengths and weaknesses of each model in handling various situations. By identifying which models perform best in specific situations, we enable the framework to make informed decisions during the ensemble phase. This ensures that actions are generated by the most suitable agent, maximizing the overall performance of the ensembled RL model.

Additionally, by continuously updating the reward distribution as new interactions occur, the framework adapts to changes in the task over time. This dynamic process ensures that the model ensemble remains effective, even as the agent encounters new and unforeseen situations. The combination of real-time situation categorization and adaptive reward tracking provides a robust foundation for the final agent selection in the ensemble phase.

\subsection{Model Ensemble}
\label{method3}
The final stage of our framework involves dynamically selecting the best-performing agent in real time based on the current task situation. This stage ensures that the actions are generated by the most suitable agent for the evolving task conditions.
During inference, we employ the same State Categorization LLM used in the previous stage to categorize the current state into one of the predefined situations every $K$ steps. This real-time situation identification enables the framework to adapt to the dynamic nature of the environment by continuously updating its understanding of the task's conditions.

For each identified situation $s$, the framework refers to the precomputed \textit{\{RewardDistribution\}}, which contains the average rewards $R_{m,s}$ achieved by each agent $m$ in situation $s$. The agent selected for generating actions in the current state is determined as:
\[
m^* = \arg\max_{m \in \mathcal{M}} R_{m,s}
\]
where $m^*$ is the selected agent, $\mathcal{M}$ is the set of all available agents, and $R_{m,s}$ is the average reward achieved by agent $m$ in situation $s$, as defined in Section~\ref{method2}. Once the model $m^*$ is selected, it generates the action for the current state based on its policy. This process is repeated throughout the task, ensuring that the ensemble dynamically adapts to changes across environments.

By utilizing the agent with the highest historical performance for each specific situation, the framework leverages the strengths of individual agents while avoiding the less-suited agents. 
\section{Experiments}
\subsection{Experimental Settings}
\label{Experimental Settings}
We assess the effectiveness of LLM-Ens using the Atari benchmark~\cite{bellemare2013arcade,kaiser2019model}. 
For the main experiments, we use LLM-Ens to ensemble different agents trained within the DQN~\cite{mnih2015human} framework as the base algorithm. 
We sample a subset of 13 tasks out of the total 26 available in Atari, focusing on tasks where the original DQN achieves stable convergence.
We set the training duration for each task between 100k and 500k steps, depending on the speed at which rewards increase during the training of the DQN algorithm.

In the LLM-Ens module, we employ GPT-4o mini\footnote{\url{https://platform.openai.com/docs/models/gpt-4o-mini}} as the core LLM and configure the key parameter in our design, the state categorization interval, to $K=30$. To ensure reproducibility, we include the detailed values of all hyper-parameters in Appendix~\ref{details}.

In terms of baselines, we compare LLM-Ens with established RL model ensemble methods, including:
\begin{itemize}
    \item \textbf{Majority Voting}~\cite{fausser2015neural,song2023ensemble}: Each agent selects an action independently, and the action with the most votes is chosen. If there is a tie, a random selection strategy is used. This method assumes that the collective decision of multiple agents leads to more robust action selection.
    \item \textbf{Rank Voting}~\cite{fausser2015neural,song2023ensemble}: Actions are ranked by each agent based on their policy preferences. The rankings are then aggregated to compute a final score for each action, and the action with the highest cumulative ranking is selected. This method ensures that all actions are considered rather than just the most frequently chosen one.
    \item \textbf{Aggregation}~\cite{li2022deep}: Each agent's action probabilities are summed equally without additional weighting. The final action is then selected based on the aggregated values. This approach maintains simplicity while leveraging collective decision-making.
    \item \textbf{Boltzmann Addition}~\cite{wiering2008ensemble}: Each agent computes action probabilities using the Boltzmann distribution, which are summed across agents before normalization. This method retains diversity in decision-making while emphasizing shared preferences among agents.
    \item \textbf{Boltzmann Multiplication}~\cite{wiering2008ensemble}: Each agent calculates action probabilities using the Boltzmann distribution. These probabilities are then multiplied across agents and renormalized, reinforcing actions that are commonly favored. The final action is sampled from this adjusted probability distribution.
\end{itemize}

\begin{table*}[ht]
\caption{Average episode rewards for ensemble methods and the best single agent across Atari environments, evaluated over 5 random seeds. The table values represent means, with parentheses indicating standard deviations. The bold and underlined fonts highlight the best and second-best results, respectively.}
\label{tab1}
\small
\centering
\resizebox{\textwidth}{!}{%
\begin{tabular}{@{}llllllll@{}}
\toprule
\multicolumn{1}{c}{\multirow{2}{*}{Environment}} & \multicolumn{1}{c}{\multirow{2}{*}{\begin{tabular}[c]{@{}c@{}}Best\\  Single\\  Agent\end{tabular}}} & \multicolumn{6}{c}{Ensemble Method} \\
\\ \cmidrule(l){3-8}
\multicolumn{1}{c}{} & \multicolumn{1}{c}{} & \begin{tabular}[c]{@{}l@{}}Majority \\ Voting\end{tabular} & \begin{tabular}[c]{@{}l@{}}Rank \\ Voting\end{tabular} & Aggregation & \begin{tabular}[c]{@{}l@{}}Boltzmann \\ Addition\end{tabular} & \begin{tabular}[c]{@{}l@{}}Boltzmann \\ Multiplication\end{tabular} & LLM-Ens (ours) \\
\midrule
BattleZone & 5200{\scriptsize(3033.15)} & 5000{\scriptsize(2549.51)} & 7000{\scriptsize(3162.28)} & {\ul{8600{\scriptsize(3507.14)}}} & 6800{\scriptsize(2949.58)} & 7000{\scriptsize(4358.9)} & \textbf{10400{\scriptsize(4159.33)}} \\
Boxing & {\ul{0{\scriptsize(3.16)}}} & -1.4{\scriptsize(6.23)} & -1.6{\scriptsize(4.51)} & {-5.6{\scriptsize(8.41)}} & -10.2{\scriptsize(6.57)} & -6.2{\scriptsize(5.76)} & \textbf{2.2{\scriptsize(3.27)}} \\
Breakout & 4.2{\scriptsize(2.17)} & 4.4{\scriptsize(1.82)} & 4.2{\scriptsize(1.48)} & 4{\scriptsize(1.73)} & {\ul{4.6{\scriptsize(0.89)}}} & 4.4{\scriptsize(0.55)} & \textbf{4.8{\scriptsize(0.84)}} \\
ChopperCommand & 740{\scriptsize(181.66)} & 680{\scriptsize(192.35)} & 760{\scriptsize(181.66)} & {\ul{860{\scriptsize(230.22)}}} & 780{\scriptsize(303.32)} & 800{\scriptsize(122.47)} & \textbf{1020{\scriptsize(216.79)}} \\
CrazyClimber & 6800{\scriptsize(1148.91)} & {\ul{7220{\scriptsize(978.26)}}} & 7000{\scriptsize(1403.57)} & {7040{\scriptsize(991.46)}} & 6320{\scriptsize(414.73)} & 7020{\scriptsize(1107.7)} & \textbf{7340{\scriptsize(1128.27)}} \\
DemonAttack & 140{\scriptsize(31.62)} & 145{\scriptsize(22.36)} & 134{\scriptsize(27.02)} & \textbf{164{\scriptsize(32.09)}} & 136{\scriptsize(23.02)} & 148{\scriptsize(30.33)} & {\ul{161{\scriptsize(51.77)}}} \\
Frostbite & 190{\scriptsize(12.25)} & 174{\scriptsize(20.74)} & {\ul{206{\scriptsize(8.94)}}} & 150{\scriptsize(50.00)} & 152{\scriptsize(43.24)} & 150{\scriptsize(36.74)} & \textbf{210{\scriptsize(36.74)}} \\
Gopher & 308{\scriptsize(133.12)} & {\ul{336{\scriptsize(82.95)}}} & 260{\scriptsize(178.89)} & 228{\scriptsize(192.15)} & 292{\scriptsize(274.08)} & 332{\scriptsize(296.85)} & \textbf{340{\scriptsize(93.81)}} \\
Krull & 1982{\scriptsize(283.41)} & 2000{\scriptsize(368.17)} & 1988{\scriptsize(421.57)} & {\ul{2032{\scriptsize(322.21)}}} & 1998{\scriptsize(146.01)} & 2028{\scriptsize(240.87)} & \textbf{2078{\scriptsize(609.57)}} \\
KungFuMaster & 5500{\scriptsize(1041.63)} & 5360{\scriptsize(907.19)} & 5420{\scriptsize(1851.22)} & 4980{\scriptsize(785.49)} & 5400{\scriptsize(927.36)} & {\ul{5580{\scriptsize(1875.37)}}} & \textbf{5600{\scriptsize(787.40)}} \\
MsPacman & {\ul{874{\scriptsize(122.6)}}} & 442{\scriptsize(135.9)} & 690{\scriptsize(451.39)} & {554{\scriptsize(221.88)}} & 494{\scriptsize(204.77)} & 738{\scriptsize(280.66)} & \textbf{1116{\scriptsize(640.37)}} \\
Qbert & 325{\scriptsize(166.77)} & 295{\scriptsize(213.16)} & 255{\scriptsize(185.74)} & 215{\scriptsize(114.02)} & \textbf{340{\scriptsize(152.68)}} & 180{\scriptsize(126.74)} & {\ul{340{\scriptsize(403.73)}}} \\
UpNDown & 1928{\scriptsize(867.28)} & 1220{\scriptsize(467.71)} & {\ul{2136{\scriptsize(901.79)}}} & \textbf{2544{\scriptsize(619.3)}} & 1980{\scriptsize(390.13)} & 1812{\scriptsize(272.8)} & 1852{\scriptsize(190.05)} \\
\bottomrule
\end{tabular}
}
\end{table*}

\begin{figure*}[ht]
    \centering
    \begin{subfigure}[t]{0.32\linewidth} 
        \includegraphics[height=4.5cm]{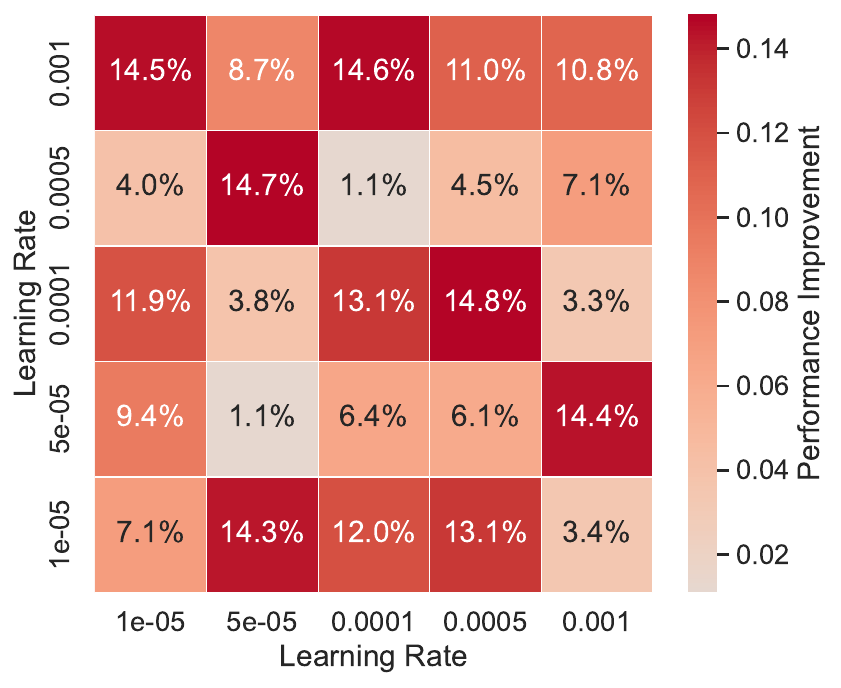} 
        \caption{\parbox{\linewidth}{\centering Learning rate experiments\\@BattleZone.}}
    \end{subfigure}
    \hfill
    \begin{subfigure}[t]{0.32\linewidth}
        \includegraphics[height=4.5cm]{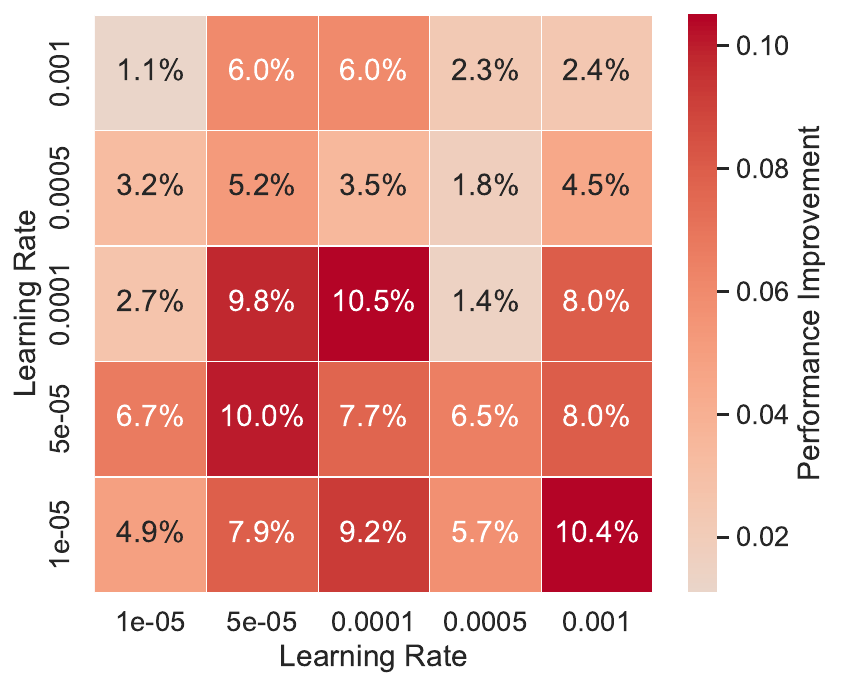}
        \caption{\parbox{\linewidth}{\centering Learning rate experiments\\@Breakout.}}
    \end{subfigure}
    \hfill
    \begin{subfigure}[t]{0.32\linewidth}
        \includegraphics[height=4.5cm]{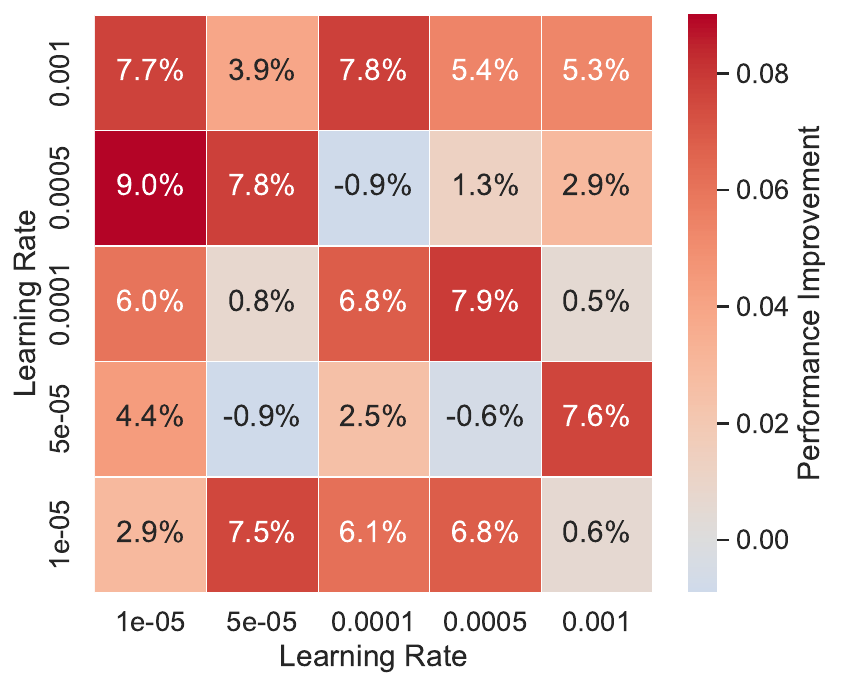}
        \caption{\parbox{\linewidth}{\centering Learning rate experiments\\@KungFuMaster.}}
    \end{subfigure}
    \vskip 0.5cm 
    \begin{subfigure}[t]{0.32\linewidth}
        \includegraphics[height=4.5cm]{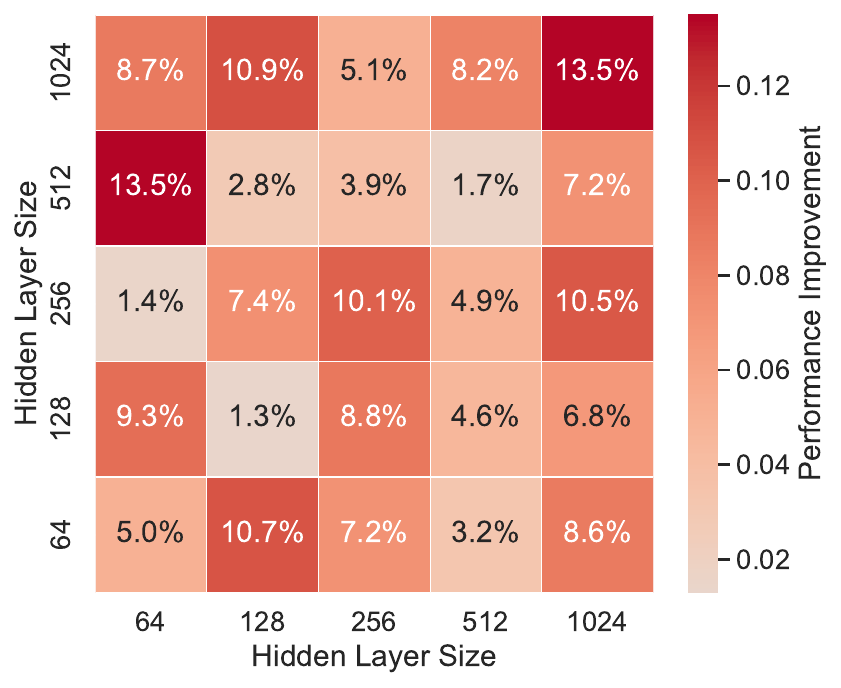}
        \caption{\parbox{\linewidth}{\centering Hidden layer size experiments\\@BattleZone.}}
    \end{subfigure}
    \hfill
    \begin{subfigure}[t]{0.32\linewidth}
        \includegraphics[height=4.5cm]{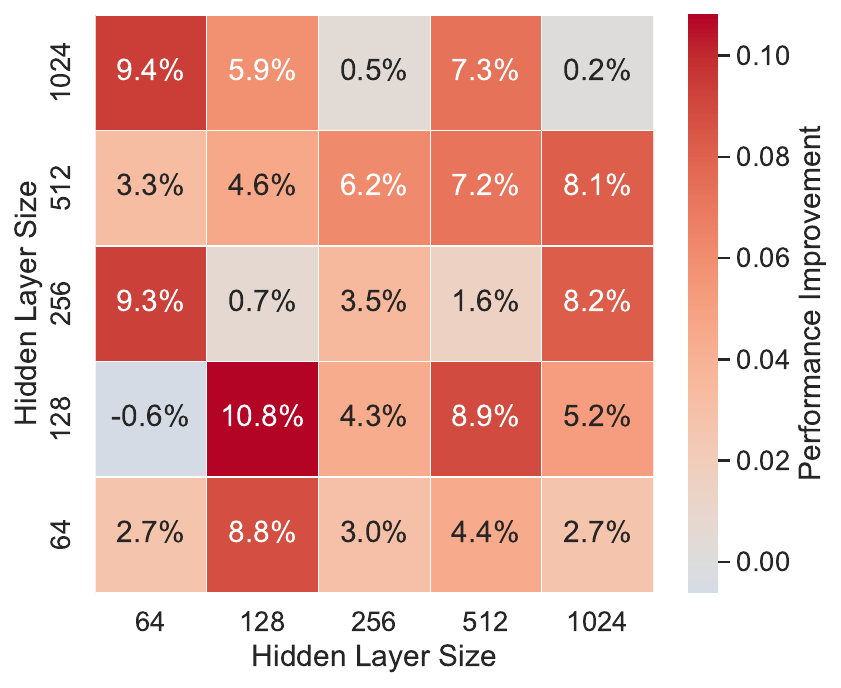}
        \caption{\parbox{\linewidth}{\centering Hidden layer size experiments\\@Breakout.}}
    \end{subfigure}
    \hfill
    \begin{subfigure}[t]{0.32\linewidth}
        \includegraphics[height=4.5cm]{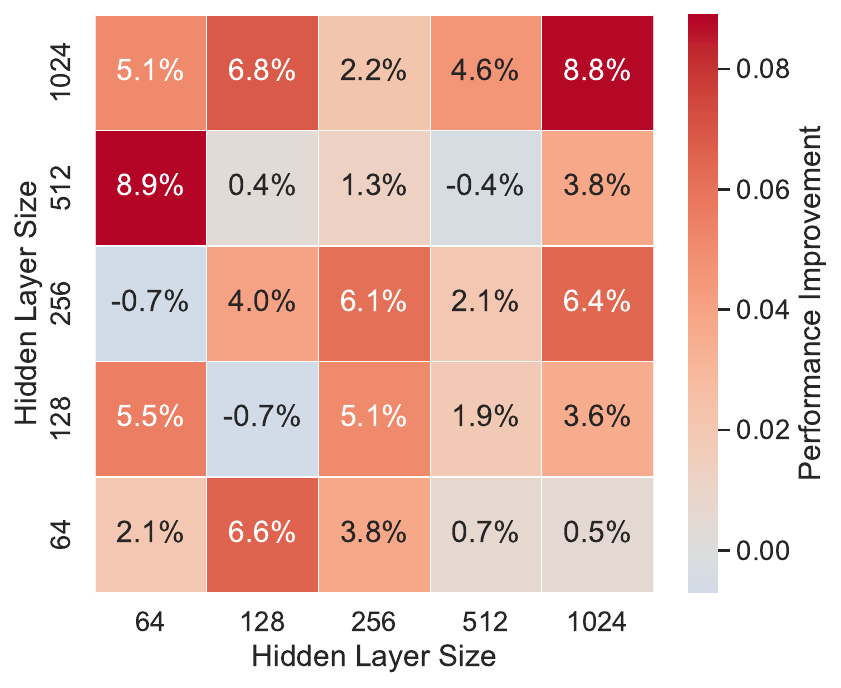}
        \caption{\parbox{\linewidth}{\centering Hidden layer size experiments\\@KungFuMaster.}}
    \end{subfigure}
    \caption{Performance improvement of LLM-Ens across different hyperparameter configurations on three Atari environments.}
    \label{fig:fig2}
\end{figure*}

\subsection{Overall Performance}
\label{performance}
To evaluate the effectiveness of LLM-Ens, we compare it with existing RL ensemble methods and the best-performing single agent across 13 Atari environments. All DQN models used for ensembling are trained with the same algorithm and hyperparameters, varying only in their random seeds. The ensemble methods include Majority Voting, Rank Voting, Aggregation, Boltzmann Addition, and Boltzmann Multiplication. The best single agent results are also provided as a baseline reference. We repeat each experiment for five times and show the results in  Table~\ref{tab1}, where the results are reported as the mean and standard deviation of the average episode rewards. Bold values indicate the best results, while underlined values indicate the second-best results.

In the results, LLM-Ens demonstrates consistent superiority across most environments, achieving significant improvements over both the best single agent and established ensemble methods.
For example, in the BattleZone environment, LLM-Ens achieves an average reward of 10400, which represents a 20.9\% improvement over the second-best method (Aggregation, 8600) and a 100\% improvement compared to the best single agent (5200).
Similarly, in ChopperCommand, LLM-Ens achieves a reward of 1020, which is 18.6\% higher than Aggregation (860) and 37.8\% higher than the best single agent (740).
And in MsPacman, LLM-Ens achieves a reward of 1116, marking a 51.2\% improvement over the second-best method (738 from Boltzmann Multiplication) and a 27.7\% improvement over the best single agent (874). These results highlight the robustness of LLM-Ens across different tasks, where traditional ensemble methods often struggle due to their inability to adapt to varying task conditions.

Among the baseline methods, Aggregation generally performs well, achieving the second-best results in several environments, such as BattleZone (8600) and UpNDown (2544). However, its performance is not always reliable. In some cases, Aggregation produces rewards even lower than the best single agent, indicating that static ensemble strategies may not always be effective. For instance, in MsPacman, Aggregation achieves a reward of 554, which is 50.4\% lower than LLM-Ens and 36.6\% lower than the best single agent. This suggests that fixed aggregation strategies, while beneficial in some environments, can lead to instability and suboptimal performance in more complex or highly dynamic tasks due to the noise from less reliable agents.
On the other hand, Boltzmann Addition and Boltzmann Multiplication, which use temperature-scaled probabilistic weighting, achieve competitive results in some environments but also exhibit inconsistencies. These methods provide slight improvements over voting-based approaches but remain limited by their static decision mechanisms, which prevent them from dynamically adapting to state-specific conditions.
Moreover, Voting-based methods, such as Majority Voting and Rank Voting, show even greater instability. Their rigid decision-making processes fail to account for the varying reliability of individual agents across different states, often leading to erratic performance. For example, in the KungFuMaster environment, Majority Voting achieves a reward of 5360, which is 4.3\% lower than the best single agent (5500), while LLM-Ens improves slightly over the best baseline (5580) with a 0.36\% increase. In MsPacman, Majority Voting produces a reward of 442, which is 60.4\% lower than LLM-Ens. These results illustrate that voting-based approaches are not well-suited for RL tasks, as they lack the adaptability required to handle diverse and evolving game conditions.

A key advantage of LLM-Ens is its adaptability across different tasks. While traditional ensemble methods exhibit fluctuations—sometimes outperforming the best single agent and sometimes falling below it—LLM-Ens consistently provides competitive or superior results. This suggests that its ability to dynamically select models based on task-specific situation classification helps mitigate the risk of suboptimal decision-making that often affects static ensemble methods.

Overall, LLM-Ens achieves substantial improvements across diverse tasks, consistently outperforming existing ensemble methods and the best single agent. By dynamically selecting the best-performing agent for each situation, LLM-Ens delivers superior performance with an improvement up to 51.2\% over baseline methods. Its ability to maintain stable and robust performance across a wide range of tasks demonstrates its potential as a powerful and adaptable ensemble approach for reinforcement learning.
Also, we showcase examples of the task-specific situations in Appendix~\ref{situations}, where the situations categorized by our Situation Generation LLM are intuitive, better illustrating the mechanism for LLM-Ens to achieve adaptability across various tasks.

\subsection{Ensemble across Different Hyper-Parameters}

Besides the capability of combining agents trained with different random seeds aforementioned, we further evaluate the effectiveness of LLM-Ens in dynamically selecting the best-performing agent with diverse hyper-parameter configurations.
We conducted experiments on three Atari tasks: Breakout, BattleZone, and KungFuMaster.
For each agent, we vary either the learning rate or the hidden layer size of the fully connected component in the DQN architecture while keeping all other parameters constant to ensure a controlled experimental setting.
Each configuration was trained with two different random seeds to assess robustness and account for stochastic variations in training outcomes.

Figure~\ref{fig:fig2} presents the experimental results in the form of heatmaps, which visually depict the performance variations across different hyperparameter settings. Each grid in the heatmap represents the performance of LLM-Ens when applied to the ensemble of agents trained with the corresponding hyperparameter values along the x- and y-axes. The color intensity in each grid cell indicates the relative improvement over the better-performing single agent, with red representing a positive improvement and blue indicating a performance drop. Figure~\ref{fig:fig2}(a–c) illustrate the results across different learning rate configurations, while Figure~\ref{fig:fig2}(d–f) correspond to variations in the hidden layer size, enabling a comprehensive evaluation of LLM-Ens's performance under different conditions.

Across all experiments, the majority of grid points are shaded red, indicating consistent performance improvements of LLM-Ens ensembled agent compared to the better-performing single agent. This result provides strong empirical evidence demonstrating the effectiveness of LLM-Ens in dynamically selecting the most suitable agent based on task-specific situations.

Among the three environments, Battlezone consistently exhibits the largest performance gains. Notably, in Figure~\ref{fig:fig2}(a), performance improvements under specific learning rate configurations reach up to 14.8 percent, highlighting LLM-Ens's strong adaptability and ability to leverage diverse models effectively in this environment where task complexity is relatively high. Similarly, Figure~\ref{fig:fig2}(d) reveals significant improvements of up to 13.5 percent with varying hidden layer sizes, further demonstrating the capability of LLM-Ens to select agents with architectural differences. In the Breakout environment, moderate but stable performance improvements are observed, with Figure~\ref{fig:fig2}(b)(e) showing gains of up to 11.7 percent and 10.8 percent, respectively. While smaller compared to Battlezone, these results still confirm LLM-Ens's effectiveness in optimizing agent selection under different hyperparameter settings.

Overall, the results confirm that LLM-Ens enables effective RL model ensemble in environments where task variability and model diversity are prominent, making it a robust and flexible approach for reinforcement learning tasks. These findings validate the proposed method's ability to dynamically adapt to task-specific conditions and hyperparameter settings, achieving significant performance gains across various configurations while maintaining a high degree of adaptability and reliability.

\section{Related Works}
\subsection{Ensemble of Reinforcement Learning Agents}
Ensemble methods in RL aim to combine multiple agents to improve decision-making and overall performance. Among these, voting~\cite{fausser2015neural,song2023ensemble,wiering2008ensemble} is a widely used strategy that determines the final action based on majority or ranking principles, where the action most frequently selected by the agents is chosen or the actions by agents are weighted by their ranks for more refined decisions.
Besides, aggregation~\cite{li2022deep} offers a simple and effective method by combining all agents’ outputs equally, while weighted aggregation improves this by assigning higher weights to models with better performance, making ensemble decisions more accurate.
Furthermore, optimal combination strategies~\cite{jalali2021oppositional} enhance the performance by selecting a subset of base models that perform best for the task at hand. This approach ensures that only the most relevant agents contribute to the final decision, avoiding potential noise from less reliable agents.
Moreover, probabilistic methods such as Boltzmann multiplication~\cite{wiering2008ensemble} assign probabilities to each action based on the Boltzmann distribution, selecting the action with the highest probability. Similarly, Boltzmann addition~\cite{wiering2008ensemble} sums the action probabilities across agents, emphasizing actions with higher overall likelihoods to make informed decisions.

Despite their widespread use, these ensemble methods rely on static strategies that cannot adapt to the dynamic nature of RL environments. They lack the flexibility to incorporate task-specific context, which limits their effectiveness in complex scenarios. In this work, we propose an approach where LLMs dynamically select and combine weak RL agents based on real-time environmental conditions, enabling a more adaptive and context-aware ensemble process.

\subsection{Enhancing RL with LLMs}

As two key areas in contemporary AI research, numerous studies have examined the integration of large language models (LLMs) to improve reinforcement learning (RL) performance~\cite{cao2024survey}. One major line of research focuses on utilizing LLMs to design reward functions that are tailored to the specifics of the environment and tasks, providing essential feedback to guide the agent's policy learning~\cite{colas2023augmenting,wu2024read,song2023self,Xie2024reward}. Another area of exploration investigates the use of LLMs to develop state representation functions, which offer more effective state inputs for the agents~\cite{wang2024llm}.
At a broader level, LLMs have been employed to decompose complex tasks into smaller sub-goals~\cite{colas2023augmenting} or to generate high-level instructions~\cite{zhou2023large}, thereby facilitating the training of RL agents. Furthermore, LLMs are increasingly being used for human-AI coordination, allowing humans to specify desired strategies for RL agents through natural language instructions~\cite{hu2023language}.
\par However, despite these advances, the application of LLMs to enhance model ensemble in RL remains an underexplored area. This paper aims to investigate and address this gap in knowledge.
\section{Conclusions}
In this paper, we propose a framework that leverages LLMs to enhance RL model ensemble.
Our approach utilizes LLMs to analyze the characteristics of different environments and assess the strengths and weaknesses of individual agents in different task situations.
This allows us to adapt the ensemble strategy to the specific characteristics of each task and adjust it dynamically during the task process, effectively combining multiple weak agents into a superior one with optimal performance.
Through extensive experiments across a variety of tasks, we demonstrate the effectiveness of our approach and highlight its compatibility with agents trained using different random seeds, hyperparameter configurations, and RL algorithms.
These results underscore the potential of our method and its wide applicability across diverse RL tasks.
\bibliographystyle{unsrt}
\bibliography{reference}

\begin{thebibliography}{10}

\bibitem{sutton2018reinforcement}
Richard~S Sutton.
\newblock Reinforcement learning: An introduction.
\newblock {\em A Bradford Book}, 2018.

\bibitem{franccois2018introduction}
Vincent Fran{\c{c}}ois-Lavet, Peter Henderson, Riashat Islam, Marc~G Bellemare, Joelle Pineau, et~al.
\newblock An introduction to deep reinforcement learning.
\newblock {\em Foundations and Trends{\textregistered} in Machine Learning}, 11(3-4):219--354, 2018.

\bibitem{silver2017mastering}
David Silver, Julian Schrittwieser, Karen Simonyan, Ioannis Antonoglou, Aja Huang, Arthur Guez, Thomas Hubert, Lucas Baker, Matthew Lai, Adrian Bolton, et~al.
\newblock Mastering the game of go without human knowledge.
\newblock {\em nature}, 550(7676):354--359, 2017.

\bibitem{vinyals2019grandmaster}
Oriol Vinyals, Igor Babuschkin, Wojciech~M Czarnecki, Micha{\"e}l Mathieu, Andrew Dudzik, Junyoung Chung, David~H Choi, Richard Powell, Timo Ewalds, Petko Georgiev, et~al.
\newblock Grandmaster level in starcraft ii using multi-agent reinforcement learning.
\newblock {\em nature}, 575(7782):350--354, 2019.

\bibitem{berner2019dota}
Christopher Berner, Greg Brockman, Brooke Chan, Vicki Cheung, Przemys{\l}aw D{\k{e}}biak, Christy Dennison, David Farhi, Quirin Fischer, Shariq Hashme, Chris Hesse, et~al.
\newblock Dota 2 with large scale deep reinforcement learning.
\newblock {\em arXiv preprint arXiv:1912.06680}, 2019.

\bibitem{ye2021mastering}
Weirui Ye, Shaohuai Liu, Thanard Kurutach, Pieter Abbeel, and Yang Gao.
\newblock Mastering atari games with limited data.
\newblock {\em Advances in neural information processing systems}, 34:25476--25488, 2021.

\bibitem{mirhoseini2021graph}
Azalia Mirhoseini, Anna Goldie, Mustafa Yazgan, Joe~Wenjie Jiang, Ebrahim Songhori, Shen Wang, Young-Joon Lee, Eric Johnson, Omkar Pathak, Azade Nazi, et~al.
\newblock A graph placement methodology for fast chip design.
\newblock {\em Nature}, 594(7862):207--212, 2021.

\bibitem{hao2021hierarchical}
Qianyue Hao, Fengli Xu, Lin Chen, Pan Hui, and Yong Li.
\newblock Hierarchical reinforcement learning for scarce medical resource allocation with imperfect information.
\newblock In {\em Proceedings of the 27th ACM SIGKDD Conference on Knowledge Discovery \& Data Mining}, pages 2955--2963, 2021.

\bibitem{hao2022reinforcement}
Qianyue Hao, Wenzhen Huang, Fengli Xu, Kun Tang, and Yong Li.
\newblock Reinforcement learning enhances the experts: Large-scale covid-19 vaccine allocation with multi-factor contact network.
\newblock In {\em Proceedings of the 28th ACM SIGKDD Conference on Knowledge Discovery and Data Mining}, pages 4684--4694, 2022.

\bibitem{hao2023gat}
Qianyue Hao, Wenzhen Huang, Tao Feng, Jian Yuan, and Yong Li.
\newblock Gat-mf: Graph attention mean field for very large scale multi-agent reinforcement learning.
\newblock In {\em Proceedings of the 29th ACM SIGKDD Conference on Knowledge Discovery and Data Mining}, pages 685--697, 2023.

\bibitem{zheng2023spatial}
Yu~Zheng, Yuming Lin, Liang Zhao, Tinghai Wu, Depeng Jin, and Yong Li.
\newblock Spatial planning of urban communities via deep reinforcement learning.
\newblock {\em Nature Computational Science}, 3(9):748--762, 2023.

\bibitem{zheng2024survey}
Yu~Zheng, Qianyue Hao, Jingwei Wang, Changzheng Gao, Jinwei Chen, Depeng Jin, and Yong Li.
\newblock A survey of machine learning for urban decision making: Applications in planning, transportation, and healthcare.
\newblock {\em ACM Computing Surveys}, 2024.

\bibitem{wang2024dyps}
Jingwei Wang, Qianyue Hao, Wenzhen Huang, Xiaochen Fan, Zhentao Tang, Bin Wang, Jianye Hao, and Yong Li.
\newblock Dyps: Dynamic parameter sharing in multi-agent reinforcement learning for spatio-temporal resource allocation.
\newblock In {\em Proceedings of the 30th ACM SIGKDD Conference on Knowledge Discovery and Data Mining}, pages 3128--3139, 2024.

\bibitem{wang2025coopride}
Jingwei Wang, Qianyue Hao, Wenzhen Huang, Xiaochen Fan, Qin Zhang, Zhentao Tang, Bin Wang, Jianye Hao, and Yong Li.
\newblock Coopride: Cooperate all grids in city-scale ride-hailing dispatching with multi-agent reinforcement learning.
\newblock In {\em Proceedings of the 31st ACM SIGKDD Conference on Knowledge Discovery and Data Mining V. 1}, pages 1457--1468, 2025.

\bibitem{adkins2024method}
Jacob Adkins, Michael Bowling, and Adam White.
\newblock A method for evaluating hyperparameter sensitivity in reinforcement learning.
\newblock {\em arXiv preprint arXiv:2412.07165}, 2024.

\bibitem{colas2018many}
C{\'e}dric Colas, Olivier Sigaud, and Pierre-Yves Oudeyer.
\newblock How many random seeds? statistical power analysis in deep reinforcement learning experiments.
\newblock {\em arXiv preprint arXiv:1806.08295}, 2018.

\bibitem{ganaie2022ensemble}
Mudasir~A Ganaie, Minghui Hu, Ashwani~Kumar Malik, Muhammad Tanveer, and Ponnuthurai~N Suganthan.
\newblock Ensemble deep learning: A review.
\newblock {\em Engineering Applications of Artificial Intelligence}, 115:105151, 2022.

\bibitem{dong2020survey}
Xibin Dong, Zhiwen Yu, Wenming Cao, Yifan Shi, and Qianli Ma.
\newblock A survey on ensemble learning.
\newblock {\em Frontiers of Computer Science}, 14:241--258, 2020.

\bibitem{song2023ensemble}
Yanjie Song, Ponnuthurai~Nagaratnam Suganthan, Witold Pedrycz, Junwei Ou, Yongming He, Yingwu Chen, and Yutong Wu.
\newblock Ensemble reinforcement learning: A survey.
\newblock {\em Applied Soft Computing}, page 110975, 2023.

\bibitem{fausser2015neural}
Stefan Fau{\ss}er and Friedhelm Schwenker.
\newblock Neural network ensembles in reinforcement learning.
\newblock {\em Neural Processing Letters}, 41:55--69, 2015.

\bibitem{perepu2020reinforcement}
Satheesh~K Perepu, Bala~Shyamala Balaji, Hemanth~Kumar Tanneru, Sudhakar Kathari, and Vivek~Shankar Pinnamaraju.
\newblock Reinforcement learning based dynamic weighing of ensemble models for time series forecasting.
\newblock {\em arXiv preprint arXiv:2008.08878}, 2020.

\bibitem{wiering2008ensemble}
Marco~A Wiering and Hado Van~Hasselt.
\newblock Ensemble algorithms in reinforcement learning.
\newblock {\em IEEE Transactions on Systems, Man, and Cybernetics, Part B (Cybernetics)}, 38(4):930--936, 2008.

\bibitem{bellemare2013arcade}
Marc~G Bellemare, Yavar Naddaf, Joel Veness, and Michael Bowling.
\newblock The arcade learning environment: An evaluation platform for general agents.
\newblock {\em Journal of Artificial Intelligence Research}, 47:253--279, 2013.

\bibitem{kaiser2019model}
Lukasz Kaiser, Mohammad Babaeizadeh, Piotr Milos, Blazej Osinski, Roy~H Campbell, Konrad Czechowski, Dumitru Erhan, Chelsea Finn, Piotr Kozakowski, Sergey Levine, et~al.
\newblock Model-based reinforcement learning for atari.
\newblock {\em arXiv preprint arXiv:1903.00374}, 2019.

\bibitem{Brown2020gpt3}
Tom~B. Brown, Benjamin Mann, Nick Ryder, Melanie Subbiah, Jared Kaplan, Prafulla Dhariwal, Arvind Neelakantan, Pranav Shyam, Girish Sastry, Amanda Askell, Sandhini Agarwal, Ariel Herbert{-}Voss, Gretchen Krueger, Tom Henighan, Rewon Child, Aditya Ramesh, Daniel~M. Ziegler, Jeffrey Wu, Clemens Winter, Christopher Hesse, Mark Chen, Eric Sigler, Mateusz Litwin, Scott Gray, Benjamin Chess, Jack Clark, Christopher Berner, Sam McCandlish, Alec Radford, Ilya Sutskever, and Dario Amodei.
\newblock Language models are few-shot learners.
\newblock In {\em NeurIPS}, 2020.

\bibitem{kalyan2023survey}
Katikapalli~Subramanyam Kalyan.
\newblock A survey of gpt-3 family large language models including chatgpt and gpt-4.
\newblock {\em Natural Language Processing Journal}, page 100048, 2023.

\bibitem{achiam2023gpt}
Josh Achiam, Steven Adler, Sandhini Agarwal, Lama Ahmad, Ilge Akkaya, Florencia~Leoni Aleman, Diogo Almeida, Janko Altenschmidt, Sam Altman, Shyamal Anadkat, et~al.
\newblock Gpt-4 technical report.
\newblock {\em arXiv preprint arXiv:2303.08774}, 2023.

\bibitem{zhong2024evaluation}
Tianyang Zhong, Zhengliang Liu, Yi~Pan, Yutong Zhang, Yifan Zhou, Shizhe Liang, Zihao Wu, Yanjun Lyu, Peng Shu, Xiaowei Yu, et~al.
\newblock Evaluation of openai o1: Opportunities and challenges of agi.
\newblock {\em arXiv preprint arXiv:2409.18486}, 2024.

\bibitem{openai2024o3}
OpenAI.
\newblock Early access for safety testing.
\newblock 2024.

\bibitem{touvron2023llama}
Hugo Touvron, Louis Martin, Kevin Stone, Peter Albert, Amjad Almahairi, Yasmine Babaei, Nikolay Bashlykov, Soumya Batra, Prajjwal Bhargava, Shruti Bhosale, et~al.
\newblock Llama 2: Open foundation and fine-tuned chat models.
\newblock {\em arXiv preprint arXiv:2307.09288}, 2023.

\bibitem{dubey2024llama}
Abhimanyu Dubey, Abhinav Jauhri, Abhinav Pandey, Abhishek Kadian, Ahmad Al-Dahle, Aiesha Letman, Akhil Mathur, Alan Schelten, Amy Yang, Angela Fan, et~al.
\newblock The llama 3 herd of models.
\newblock {\em arXiv preprint arXiv:2407.21783}, 2024.

\bibitem{chowdhery2023palm}
Aakanksha Chowdhery, Sharan Narang, Jacob Devlin, Maarten Bosma, Gaurav Mishra, Adam Roberts, Paul Barham, Hyung~Won Chung, Charles Sutton, Sebastian Gehrmann, et~al.
\newblock Palm: Scaling language modeling with pathways.
\newblock {\em Journal of Machine Learning Research}, 24(240):1--113, 2023.

\bibitem{hao2024hlm}
Qianyue Hao, Jingyang Fan, Fengli Xu, Jian Yuan, and Yong Li.
\newblock Hlm-cite: Hybrid language model workflow for text-based scientific citation prediction.
\newblock {\em arXiv preprint arXiv:2410.09112}, 2024.

\bibitem{lan2024stance}
Xiaochong Lan, Chen Gao, Depeng Jin, and Yong Li.
\newblock Stance detection with collaborative role-infused llm-based agents.
\newblock In {\em Proceedings of the international AAAI conference on web and social media}, volume~18, pages 891--903, 2024.

\bibitem{zhao2023survey}
Wayne~Xin Zhao, Kun Zhou, Junyi Li, Tianyi Tang, Xiaolei Wang, Yupeng Hou, Yingqian Min, Beichen Zhang, Junjie Zhang, Zican Dong, et~al.
\newblock A survey of large language models.
\newblock {\em arXiv preprint arXiv:2303.18223}, 2023.

\bibitem{chang2024survey}
Yupeng Chang, Xu~Wang, Jindong Wang, Yuan Wu, Linyi Yang, Kaijie Zhu, Hao Chen, Xiaoyuan Yi, Cunxiang Wang, Yidong Wang, et~al.
\newblock A survey on evaluation of large language models.
\newblock {\em ACM Transactions on Intelligent Systems and Technology}, 15(3):1--45, 2024.

\bibitem{xu2025towards}
Fengli Xu, Qianyue Hao, Zefang Zong, Jingwei Wang, Yunke Zhang, Jingyi Wang, Xiaochong Lan, Jiahui Gong, Tianjian Ouyang, Fanjin Meng, et~al.
\newblock Towards large reasoning models: A survey of reinforced reasoning with large language models.
\newblock {\em arXiv preprint arXiv:2501.09686}, 2025.

\bibitem{mnih2015human}
Volodymyr Mnih, Koray Kavukcuoglu, David Silver, Andrei~A Rusu, Joel Veness, Marc~G Bellemare, Alex Graves, Martin Riedmiller, Andreas~K Fidjeland, Georg Ostrovski, et~al.
\newblock Human-level control through deep reinforcement learning.
\newblock {\em Nature}, 518(7540):529--533, 2015.

\bibitem{li2022deep}
Guohui Li, Ming Dong, Lingfeng Ming, Changyin Luo, Han Yu, Xiaofei Hu, and Bolong Zheng.
\newblock Deep reinforcement learning based ensemble model for rumor tracking.
\newblock {\em Information Systems}, 103:101772, 2022.

\bibitem{jalali2021oppositional}
Seyed Mohammad~Jafar Jalali, Milad Ahmadian, Sajad Ahmadian, Abbas Khosravi, Mamoun Alazab, and Saeid Nahavandi.
\newblock An oppositional-cauchy based gsk evolutionary algorithm with a novel deep ensemble reinforcement learning strategy for covid-19 diagnosis.
\newblock {\em Applied Soft Computing}, 111:107675, 2021.

\bibitem{cao2024survey}
Yuji Cao, Huan Zhao, Yuheng Cheng, Ting Shu, Guolong Liu, Gaoqi Liang, Junhua Zhao, and Yun Li.
\newblock Survey on large language model-enhanced reinforcement learning: Concept, taxonomy, and methods.
\newblock {\em arXiv preprint arXiv:2404.00282}, 2024.

\bibitem{colas2023augmenting}
C{\'e}dric Colas, Laetitia Teodorescu, Pierre-Yves Oudeyer, Xingdi Yuan, and Marc-Alexandre C{\^o}t{\'e}.
\newblock Augmenting autotelic agents with large language models.
\newblock In {\em Conference on Lifelong Learning Agents}, pages 205--226. PMLR, 2023.

\bibitem{wu2024read}
Yue Wu, Yewen Fan, Paul~Pu Liang, Amos Azaria, Yuanzhi Li, and Tom~M Mitchell.
\newblock Read and reap the rewards: Learning to play atari with the help of instruction manuals.
\newblock {\em Advances in Neural Information Processing Systems}, 36, 2024.

\bibitem{song2023self}
Jiayang Song, Zhehua Zhou, Jiawei Liu, Chunrong Fang, Zhan Shu, and Lei Ma.
\newblock Self-refined large language model as automated reward function designer for deep reinforcement learning in robotics.
\newblock {\em arXiv preprint arXiv:2309.06687}, 2023.

\bibitem{Xie2024reward}
Tianbao Xie, Siheng Zhao, Chen~Henry Wu, Yitao Liu, Qian Luo, Victor Zhong, Yanchao Yang, and Tao Yu.
\newblock Text2reward: Reward shaping with language models for reinforcement learning.
\newblock In {\em {ICLR}}. OpenReview.net, 2024.

\bibitem{wang2024llm}
Boyuan Wang, Yun Qu, Yuhang Jiang, Jianzhun Shao, Chang Liu, Wenming Yang, and Xiangyang Ji.
\newblock Llm-empowered state representation for reinforcement learning.
\newblock {\em arXiv preprint arXiv:2407.13237}, 2024.

\bibitem{zhou2023large}
Zihao Zhou, Bin Hu, Chenyang Zhao, Pu~Zhang, and Bin Liu.
\newblock Large language model as a policy teacher for training reinforcement learning agents.
\newblock {\em arXiv preprint arXiv:2311.13373}, 2023.

\bibitem{hu2023language}
Hengyuan Hu and Dorsa Sadigh.
\newblock Language instructed reinforcement learning for human-ai coordination.
\newblock In {\em International Conference on Machine Learning}, pages 13584--13598. PMLR, 2023.

\end{thebibliography}
\newpage
\onecolumn
\appendix
\section{Implementation Details}
\label{details}
In this section, we provide the main implementation details for reproducibility in Table~\ref{Table6}.
Please refer to our source code at \url{https://anonymous.4open.science/r/LLM4RLensemble-F7EE} for the exact usage of each hyperparameters and more details.

\begin{table*}[h]
\caption{Implementation details.}
\label{Table6}
\centering
\begin{tabular}{@{}ccc@{}}
\toprule
Module                    & Element              & Detail                                  \\ \midrule
\multirow{4}{*}{System}   & OS                   & Ubuntu 22.04.2                          \\
                          & CUDA                 & 11.7                                    \\
                          & Python               & 3.11.4                                  \\
                          & Device               & 8*NVIDIA A100 80G                       \\
\midrule
\multirow{9}{*}{DQN}
& $\gamma$   & 0.99  \\ 
& Batch Size &  256 \\ 
& Interval of target network updating & 1000 \\
& Optimizer   & Adam  \\ 
& Learning rate & 0.00005 \\
& Replay buffer size & 10000 \\
& Start epsilon & 1 \\
& Min epsilon & 0.1 \\
& Epsilon decay per step  & 0.99999 \\

\midrule
\multirow{2}{*}{Situation Generation LLM} & Model name           & gpt-4o-mini \\
                          & Temperature          & 1.0                                     \\ \midrule
\multirow{2}{*}{State Categorization LLM} & Model name           & gpt-4o-mini \\
                          & Temperature          & 1.0                                    \\
\bottomrule
\end{tabular}
\end{table*}

\newpage
\section{Examples of Task-specific Situations}
\label{situations}
\begin{figure*}[ht]
    \centering
    \begin{subfigure}[b]{0.45\linewidth} 
        \centering
        \includegraphics[width=\linewidth]{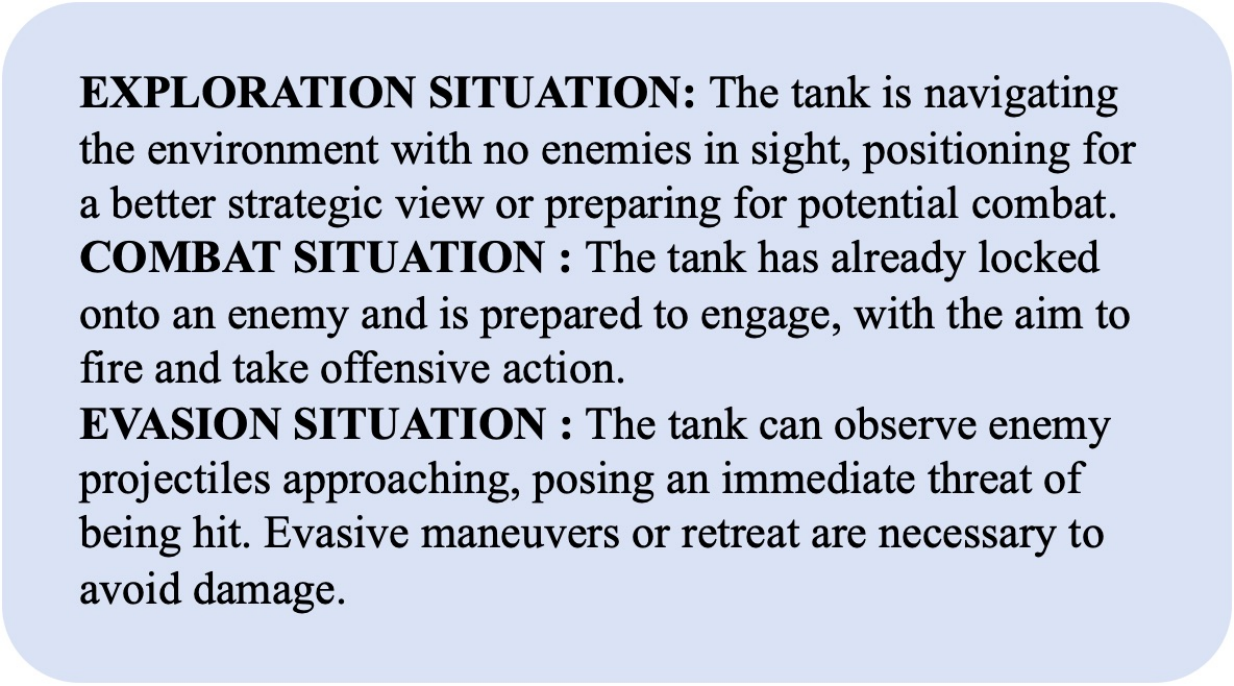}
        \caption{Situations@BattleZone.}
    \end{subfigure}
    \hspace{0.05\linewidth} 
    \begin{subfigure}[b]{0.45\linewidth}
        \centering
        \includegraphics[width=\linewidth]{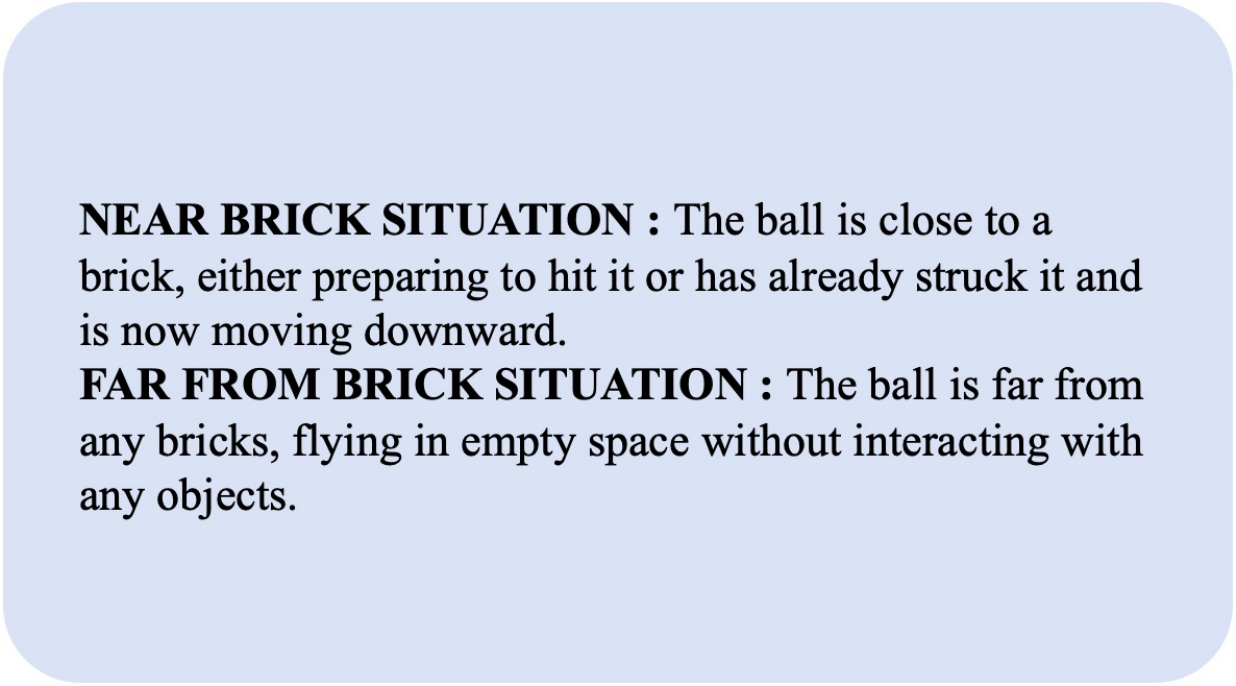}
        \caption{Situations@Breakout.}
    \end{subfigure}

    \vskip 0.5cm 
    \begin{subfigure}[b]{0.45\linewidth}
        \centering
        \includegraphics[width=\linewidth]{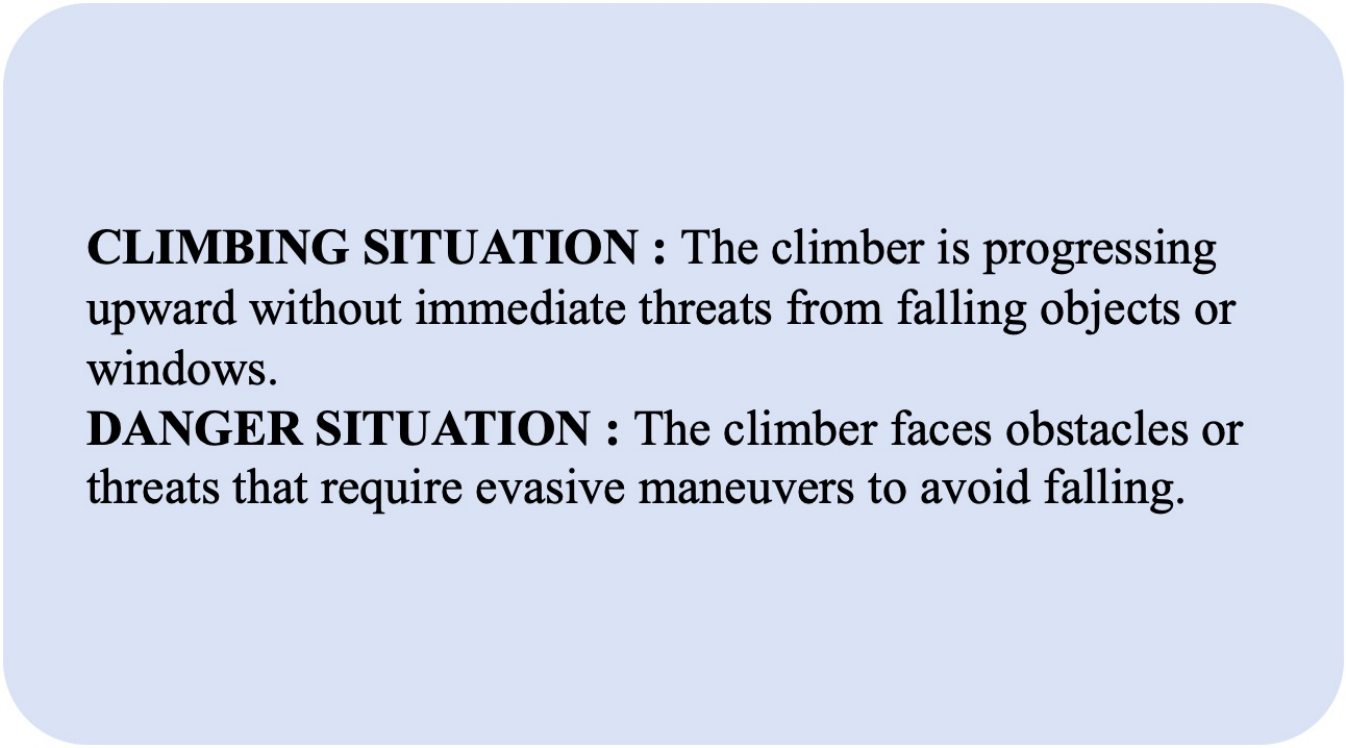}
        \caption{Situations@CrazyClimber.}
    \end{subfigure}
    \hspace{0.05\linewidth} 
    \begin{subfigure}[b]{0.45\linewidth}
        \centering
        \includegraphics[width=\linewidth]{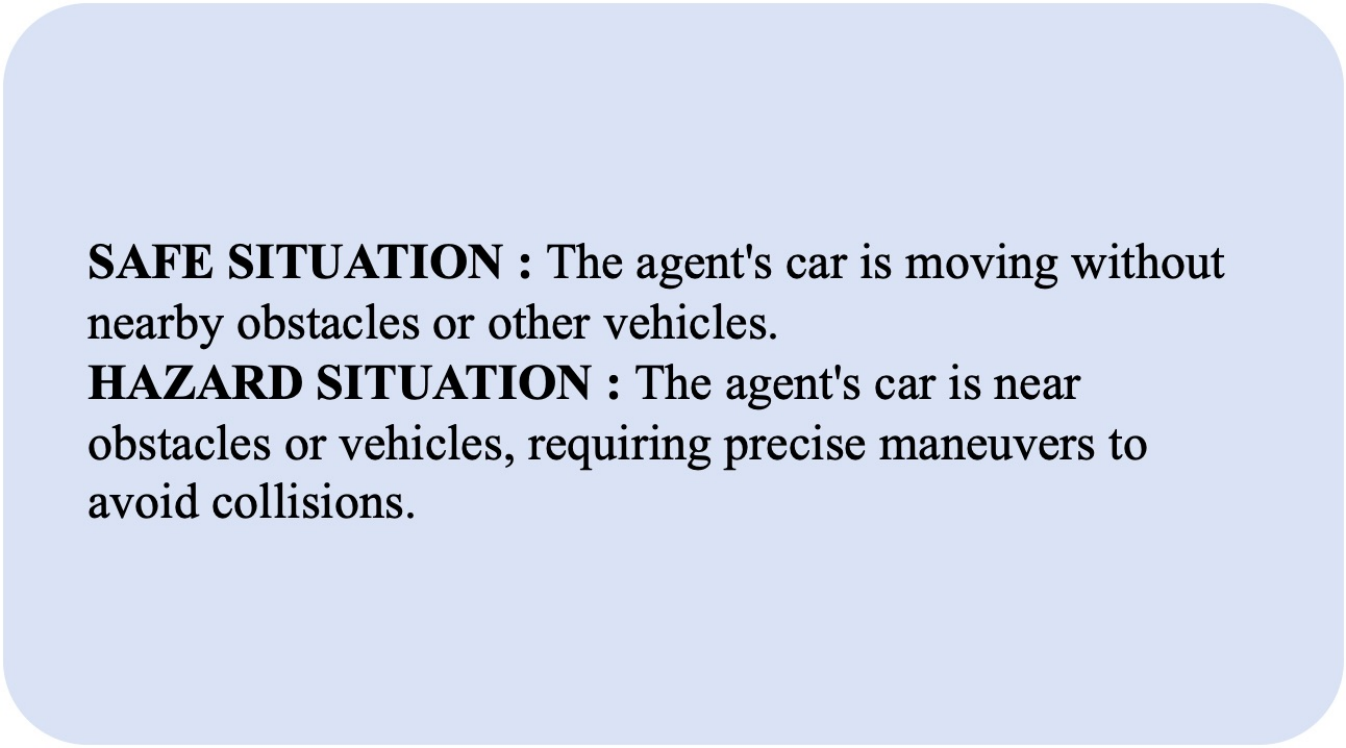}
        \caption{Situations@UpNDown.}
    \end{subfigure}

    \caption{Examples of situations categorized by the Situation Generation LLM in different environments.}
    \label{fig:example}
\end{figure*}

To illustrate how the Situation Generation LLM categorizes states into different situations, we provide concrete examples of the generated situations in Figure~\ref{fig:example}.

These examples demonstrate the LLM’s ability to capture key environmental features and group similar states under a unified classification.
By leveraging the task description and state observations, the LLM formulates a structured representation of the task dynamics, ensuring consistency in downstream decision-making. 
The generated situations not only facilitate model selection in the ensemble phase but also improve interpretability by summarizing complex state variations into distinct categories.

\begin{figure*}[ht]
    \centering
    \begin{subfigure}[b]{0.20\linewidth} 
        \centering
        \includegraphics[width=\linewidth]{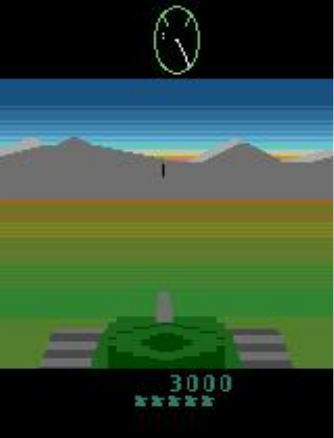}
        \caption{Exploration situation.}
    \end{subfigure}
    \hspace{0.03\linewidth} 
    \begin{subfigure}[b]{0.20\linewidth}
        \centering
        \includegraphics[width=\linewidth]{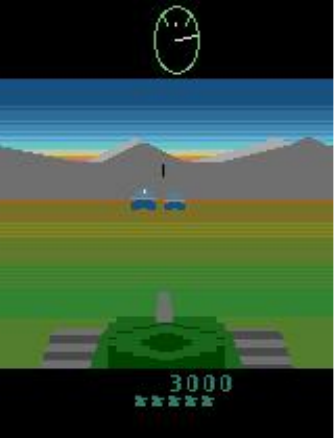}
        \caption{Combat situation.}
    \end{subfigure}
    \hspace{0.03\linewidth} 
    \begin{subfigure}[b]{0.20\linewidth}
        \centering
        \includegraphics[width=\linewidth]{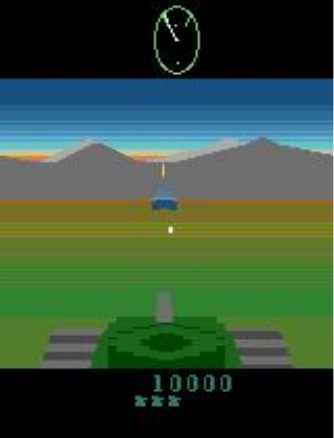}
        \caption{Evasion situation.}
    \end{subfigure}

    \caption{Representative frames for each situation categorized by the Situation Generation LLM in the BattleZone environment.}
    \label{fig:frame}
\end{figure*}

To further illustrate the validity of the generated situations, we provide frame examples corresponding to each categorized situation in the BattleZone environment in Figure~\ref{fig:frame}. These visual examples demonstrate how the Situation Generation LLM effectively distinguishes between different task-specific contexts based on environmental cues. For instance, the “Exploration Situation” captures the tank navigating the terrain without immediate threats, positioning itself strategically for future actions. The “Combat Situation” arises when the tank locks onto an enemy and prepares for an offensive strike, indicating an active engagement phase. Finally, the “Evasion Situation” depicts the tank detecting incoming projectiles, prompting the agent to execute evasive maneuvers to avoid damage. By mapping raw state observations to well-defined situations, the LLM enables a high-level understanding of the agent’s interactions within the environment.

\newpage
\section{Discussion}
\subsection{Limitation}
\label{Appendix:limitation}
One major limitation of our method lies in LLMs' illusion problem.
Despite average performance improvement, LLMs may output unfaithful analysis under certain circumstances and poison specific training processes.
When applied to real-world applications, these training trails may cause negative outcome.
Therefore, how to verify the output of LLM agents and improve the reliability of our workflow worth future studies.

\subsection{Code of ethics}
\label{Appendix:CoE}
This study uses fully open-source or publicly available models and benchmarks, adhering to their respective licenses. All resources are properly cited in Sections~\ref{Experimental Settings}. The selected benchmarks and models are well-established, representative, and free from bias or discrimination.

\subsection{Broader impacts}
\label{Appendix:Broader_impacts}
Our method holds significant potential to influence the broader domain of large language models (LLMs) and reinforcement learning (RL), a cross-research area that continues to attract substantial attention. Beyond the specific tasks demonstrated in our experiments, our approach is adaptable to a wider array of complex problems. Besides, its underlying design principles could inspire further research into leveraging LLMs to enhance various facets of RL algorithms—from policy representation and exploration strategies to reward shaping—ultimately fostering the development of more robust and intelligent AI systems.

\end{document}